\documentclass{egpubl}
\usepackage{eg2022}
 
\ConferencePaper        
\usepackage[T1]{fontenc}
\usepackage{dfadobe}  

\usepackage{cite}  
\BibtexOrBiblatex
\electronicVersion
\PrintedOrElectronic
\ifpdf \usepackage[pdftex]{graphicx} \pdfcompresslevel=9
\else \usepackage[dvips]{graphicx} \fi

\usepackage{egweblnk} 

\usepackage{cuted}
\usepackage{capt-of}
\usepackage{pifont}
\usepackage{graphics}
\usepackage{amsmath}
\usepackage{soul}
\usepackage{makecell}

\title{Real-time Virtual-Try-On from a Single Example Image through Deep Inverse Graphics and Learned Differentiable Renderers}

\author[R. Kips et al.]
{\parbox{\textwidth}{\centering R. Kips $^{1,2}$\orcid{0000-0003-0231-6415}
        , R. Jiang$^{3}$\orcid{0000-0001-6770-1379} 
        , S. Ba$^{1}$ \orcid{0000-0003-4581-9880}
        , B. Duke$^{3}$\orcid{0000-0001-6770-1379} 
        , M. Perrot$^{1}$\orcid{0000-0003-3456-8529} 
        , P. Gori$^{2}$\orcid{0000-0003-3456-8529}
        and I. Bloch$^{2,4}$\orcid{0000-0002-6984-1532} 
        }
        \\
{\parbox{\textwidth}{\centering $^1$L'Or\'eal Research and Innovation, France\\
         $^2$LTCI, T\'el\'ecom Paris, Institut Polytechnique de Paris, France \\
         $^3$ Modiface, Canada
        $^4$ Sorbonne Universit\'e, CNRS, LIP6, France\\
       }
}
}

\begin{document}


\maketitle
\begin{abstract}
Augmented reality applications have rapidly spread across online retail platforms and social media, allowing consumers to virtually try-on a large variety of products, such as makeup, hair dying, or shoes.
However, parametrizing a renderer to synthesize realistic images of a given product remains a challenging task that requires expert knowledge. 
While recent work has introduced neural rendering methods for virtual try-on from example images, current approaches are based on large generative models that cannot be used in real-time on mobile devices. This calls for a hybrid method that combines the advantages of computer graphics and neural rendering approaches. 
In this paper, we propose a novel framework based on deep learning to build a real-time inverse graphics encoder that learns to map a single example image into the parameter space of a given augmented reality rendering engine.
Our method leverages self-supervised learning and does not require labeled training data, which makes it extendable to many virtual try-on applications. 
Furthermore, most augmented reality renderers are not differentiable in practice due to algorithmic choices or implementation constraints to reach real-time on portable devices. 
To relax the need for a graphics-based differentiable renderer in inverse graphics problems, we introduce a trainable imitator module. Our imitator is a generative network that learns to accurately reproduce the behavior of a given non-differentiable renderer. We propose a novel rendering sensitivity loss to train the imitator, which ensures that the network learns an accurate and continuous representation for each rendering parameter. 
Automatically learning a differentiable renderer, as proposed here, could be beneficial for various inverse graphics tasks.
Our framework enables novel applications where consumers can virtually try-on a novel unknown product from an inspirational reference image on social media.
It can also be used by computer graphics artists to automatically create realistic rendering from a reference product image.

\ccsdesc[500]{Computing methodologies~Computer vision}
\ccsdesc[500]{Computing methodologies~Machine learning}
\ccsdesc[500]{Computing methodologies~Computer graphics}

\printccsdesc   
\end{abstract}  

\begin{figure*}[!]
\begin{center}
\includegraphics[width=0.8\textwidth]{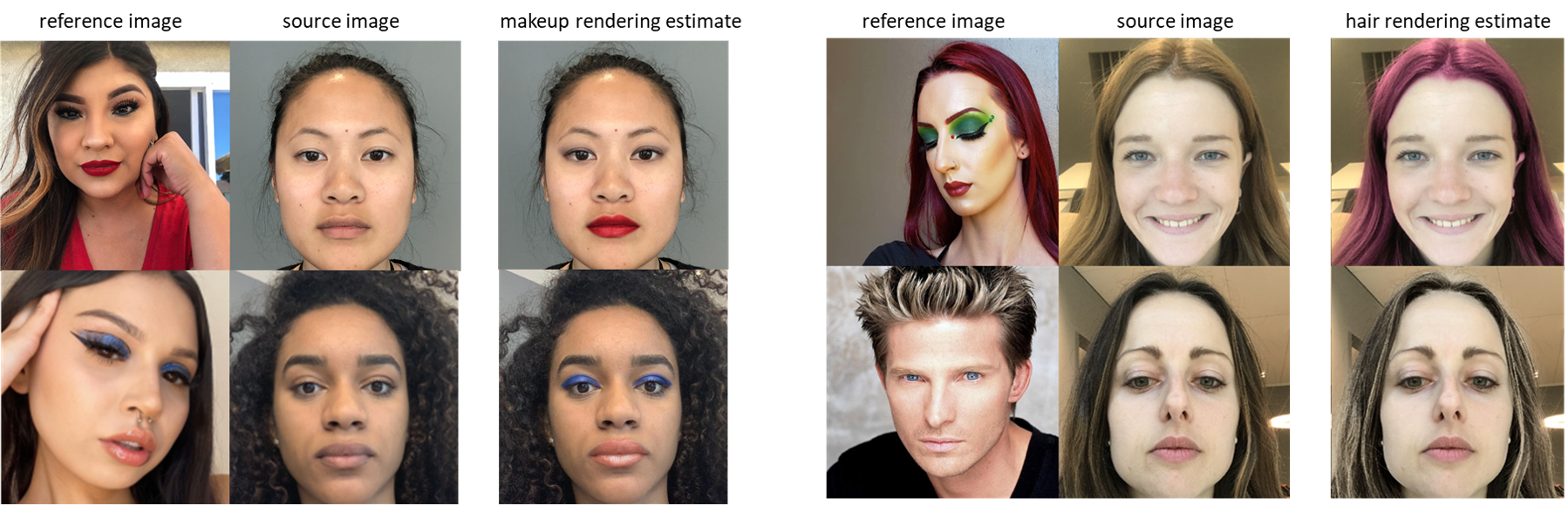}
\end{center}
    \caption{Our hybrid framework uses a deep learning based inverse graphics encoder that learns to map an example reference image into the parameter space of a computer graphics renderer. The renderer can then be used to render a virtual try-on in real-time on mobile devices.  We illustrate the performance of our framework on makeup (lipstick and eye shadow) and hair color virtual try-on.}
\label{fig:header_figure}
\end{figure*}

\section{Introduction}

The increasing development of digital sales and online shopping has been accompanied by the development of various technologies to improve the digital customer experience. 
In particular, augmented reality (AR) has rapidly spread across online retail platforms and social media, allowing consumers to virtually try-on a large variety of products like makeup, hair dying, glasses, or shoes. To reach a realistic experience, such augmented reality renderers are expected to run in real-time with the limited resources of portable devices. 
Generally, commercial virtual try-on applications need to render not a single product, but an entire range of items with various appearances. For this reason, AR renderers are commonly parameterized by a set of graphics parameters that control the variation of appearance across the various products to render. 
In practice, setting these parameters to obtain realistic rendering for hundreds of products in a digital store is a tedious task that requires expert knowledge in computer graphics. 

The rapidly emerging field of \textit{inverse graphics} provides various solutions for estimating graphics parameters from natural images using differentiable rendering \cite{kato2020differentiable}. 
Most AR renderers that run in real-time on portable devices are not differentiable in practice due to algorithmic choices or implementation constraints \cite{Gomes2012ComputerGraphics}.
Indeed, replacing non-differentiable operations with their differentiable approximations leads to less optimized computation speed, and would potentially lead to large re-implementation costs to ensure compatibility on multiple platforms.
This makes conventional inverse graphics solutions non-suitable for our problem. 

Recently, a novel family of methods based on \textit{neural rendering} has introduced the task of \textit{image based virtual try-on}, which brings new perspectives for this problem \cite{Tewari2020SOA-NeuralRendering}. 
This task consists in extracting a product appearance from a single reference image and synthesizing it on the image of another person.
However, existing methods in this domain are often based on large generative networks that suffer from temporal inconsistencies, and cannot be used to process a video stream in real-time on mobile devices \cite{kips2021inversemakeup, li2018beautygan}. 

To address this issue we propose a hybrid solution that leverages the speed and portability of computer graphics based renderers, together with the appearance extraction capabilities of neural-based methods.
Our contributions can be summarized as follow :
\begin{itemize}
    \item To relax the need for a graphics-based differentiable renderer in inverse graphics problems, we introduce a trainable imitator module. Our imitator is a generative network that learns to accurately reproduce the behavior of a given non-differentiable renderer. To train the imitator, we propose a novel rendering sensitivity loss which ensures
    that the network learns an accurate and continuous representation for each rendering parameter. This method for automatically learning a differentiable renderer could be beneficial for various inverse graphics tasks.
    \item We introduce a novel framework for image-based virtual try-on, using an inverse graphics encoder module that learns to map a single example image into the space of parameters of a rendering engine. This model is trained using an imitator network and a self-supervised approach which does not require labeled training data.
    \item We assess the effectiveness of our approach by investigating two well-established problems: makeup and hair color virtual try-on (see Figure~\ref{fig:header_figure}). These problems are based on very different rendering principles, respectively physically-based computer graphics, and pixel statistics manipulation. 
 \end{itemize}
 
Our method enables new applications where consumers can virtually try-on a novel unknown product from a reference inspirational image on social media. It can also be used by computer graphics artists to automatically create realistic renderings from a reference product image.
We believe that our framework can be easily adapted to other augmented reality applications since it only requires the availability of a conventional parametrized renderer.  

\section{Related works}

\begin{figure*}[t!]
\begin{center}
   \includegraphics[width=0.9\linewidth]{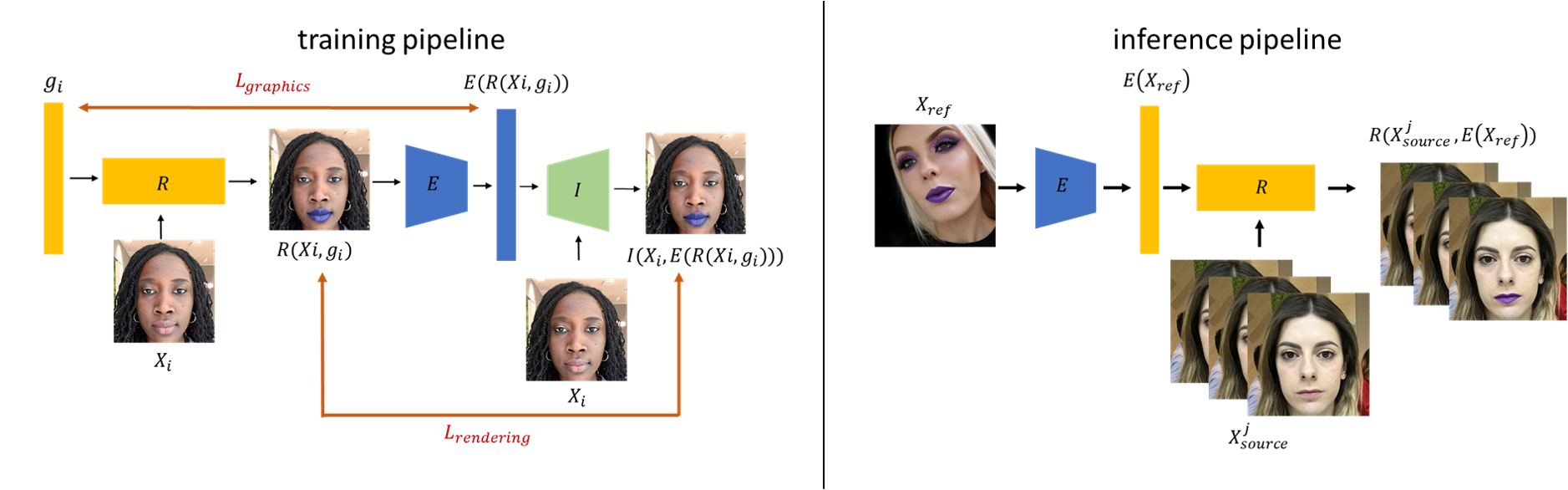}
\end{center}
  \caption{Our inference and training pipeline. The inverse graphics encoder $E$ is trained using a self-supervised approach based on the random sampling of graphics vectors $g$ sent to a renderer $R$ to generate training data. For each synthetic image, a graphics loss function $L_{graphics}$ enforces $E$ to estimate the corresponding rendering parameters. To ensure additional supervision in the image space, our learned imitator module is used to compute a differentiable rendering estimate from the encoder output. A rendering loss function $L_{rendering}$ is then computed using a perceptual distance between the rendered image and its reconstructed estimate. For inference, the imitator is discarded and we use the original renderer to reach real-time performances.
  }
\label{fig:encoder_pipeline}
\end{figure*}

\subsection{Inverse Graphics and Differentiable Rendering}

Given a natural image, \textit{inverse graphics} approaches aim to estimate features that are typically used in computer graphics scene representation, such as HDR environment map~\cite{song2019neural, somanath2020hdr} or meshes of 3D objects such as faces~\cite{FLAME:SiggraphAsia2017}. The idea of using a neural network for learning to map images to rendering parameters was first introduced in \cite{kulkarni2015deep} and is now widely spread.
Similarly to our problem, some applications focus on solutions to assist computer graphics artists to accelerate their work, such as spatially-varying bidirectional reflectance fields (SVBRDF) estimation from smartphone flash images~\cite{Deschaintre2018, Henzler2021}. 

Most inverse graphics methods build on the rapidly growing field of \textit{differentiable rendering}. This area focuses on developing differentiable operations for replacing the non-differentiable modules of computer graphics rendering pipelines \cite{Liraytracer2018, Laine2020}. Nowadays,
differentiable renderers are key components for solving inverse graphics problems. 
Indeed, the vanilla inverse graphics approach consists in estimating the set of rendering parameters by minimizing photometric or perceptual distances \cite{wang2003multiscale, zhang2018unreasonable} between a reference image and the synthesized image using stochastic gradient descent, as done in \cite{Laine2020}.
However, in addition to the requirement of a differentiable renderer, such an approach is slow since a gradient descent needs to be computed at inference for each new reference image. Thus, these methods are not suited to real-time applications. 
More recent methods, such as \cite{Henzler2021}, avoid using gradient descent during inference 
by training an end-to-end neural network, while a differentiable renderer is used to provide feedback during training. This approach results in faster inference since only a single forward pass of the model is needed. In this paper, we propose to build upon this approach.
Finally, another method consists in training graphics parameters estimators by using datasets composed of physical ground truth measurements, such as HDR maps \cite{somanath2020hdr} or OLAT images \cite{Pandey2021}. Even though these approaches led to successful applications, they are difficult to generalize to many different inverse rendering problems as ground truth measurements are extremely costly to acquire at a large scale. 

\subsection{Augmented Reality Renderers}
Augmented reality renderers are a particular category of computer graphics pipelines where the objective is to realistically synthesize an object in an image or video of a real scene. In general, AR renderers are composed of one or several scene perception modules whose role is to estimate relevant scene information from the source image, that is then passed to a rendering module. 
For instance, many portraits based AR applications are based on facial landmarks estimation \cite{kazemi2014one} used to compute the position of a synthetic object such as glasses that are then blended on the face image \cite{azevedo2016augmented}. 
Similarly, other popular scene perception methods for augmented reality focus on hand tracking \cite{wang2020rgb2hands, Zhang2020}, body pose estimation \cite{Bazarevsky2020}, hair segmentation \cite{levinshtein2018real}, or scene depth estimation \cite{kopf2020robust}.
Furthermore, most augmented reality applications target video-based problems and deployment on mobile devices. For this reason, reaching real-time with limited computation resources is usually an important focus of research in this field, as illustrated in \cite{48501, Bazarevsky2020, kopf2020robust, Li2019}. 
Many AR applications focus on virtual try-on tasks in order to enhance the consumer experience in digital stores. Popular applications introduce virtual try-on for lipstick \cite{Sokal}, hair color \cite{Tkachenka}, or nail polish \cite{Duke2019}, reaching realistic results in real-time on mobile devices.
However, for such methods, the renderer needs to be manually parametrized by an artist to obtain a realistic rendering of a targeted product. Users are restricted to select a product within a pre-defined range, and cannot virtually try a novel product from a given new reference image.

\subsection{Image-Based Virtual Try-On}

While conventional AR renderers render objects that are previously created by computer graphics artists, a recent research direction based on neural rendering has proposed the novel task of \textit{image-based virtual try-on}. The objective consists in extracting a product appearance from a given reference image and realistically synthesizing this product in the image of another person. 
Most methods in this field are built on a similar approach, that proposes to use a neural network to extract product features into a latent space. Most of the methods in this field use a similar approach by training a neural network that encodes the target product features into a latent space. Then, this product representation is decoded and rendered on the source image using generative models such as generative advesarial networks (GANs) \cite{goodfellow2020generative} or variational auto-encoders (VAEs) \cite{kingma2014auto}. 
In particular, this idea has been successfully used for makeup transfer \cite{li2018beautygan, Jiang_2020_CVPR}, hair synthesis \cite{saha2021loho, kim2021k} and is rapidly emerging in the field of fashion articles \cite{jetchev2017conditional}. Recent methods attempt to provide controllable rendering \cite{kips2020gan}, or propose to leverage additional scene information in their models, such as segmentation masks for fashion items \cite{choi2021viton, ge2021disentangled} or UV maps for makeup \cite{m_Nguyen-etal-CVPR21}. 
Another category of model leverages user's  input instead of example image to control the image synthesis. In particular, for hair virtual try-on the model from  \cite{xiao2021sketchhairsalon} uses sketch input to explicitly control the hairstyle, while StyleCLIP \cite{Patashnik_2021_ICCV} uses text input to control hair and makeup attributes in an image.

However, current virtual try-on from example image methods suffer from several limitations. First, these neural rendering methods are based on large generative models that cannot be used to produce high-resolution renderings in real-time on mobile devices. 
Furthermore, such models often lead to poor results when used on videos, since generative models are known to produce time inconsistencies artifacts. Even though recent works attempt to address this issue by training post-processing models \cite{chu2020learning, thimonier2021styletemp}, they cannot be used in real-time.
More recently, an inverse graphics approach has been introduced for makeup~\cite{kips2021inversemakeup}, bringing interesting perspectives for reaching real-time image-based virtual try-on. In this paper, we propose to build on this approach to propose a more robust and general framework. 

\section{Framework Overview}

\begin{figure}[!]
\begin{center}
   \includegraphics[width=1.0\linewidth]{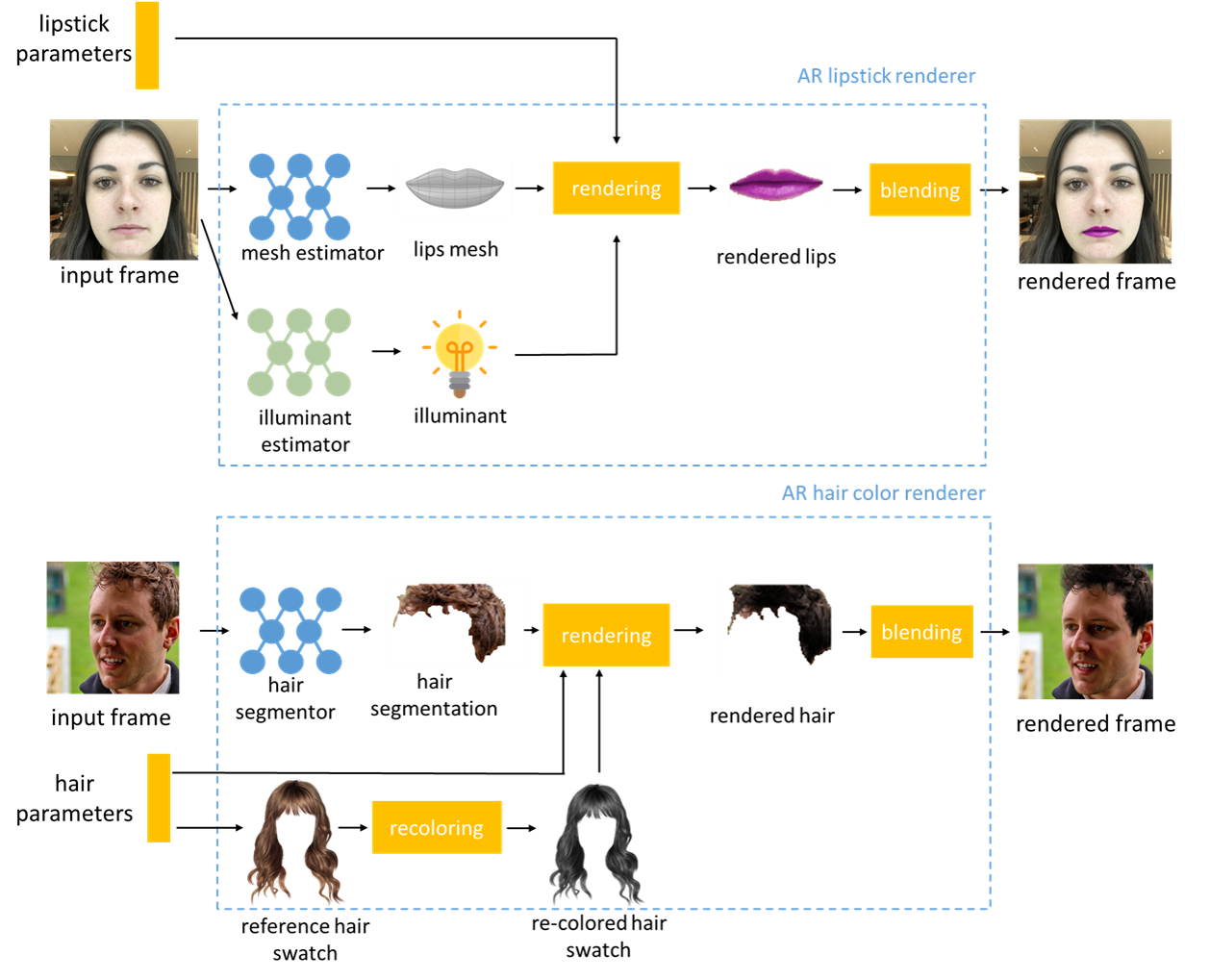}
\end{center}
   \caption{Description of our makeup and hair color augmented reality renderers. They are composed of scene perception modules that compute scene information which is then passed to a computer graphics renderer $R$. The lipstick renderer uses physically-based rendering while the hair color renderer uses pixel statistics manipulations. 
   For both renderers, the appearance of the product to render is controlled by a vector of graphics parameters $g$.
   The eye shadow and lipstick renderer are based on similar principles and parameters.}
\label{fig:ar_renderers}
\end{figure}

In this section, we present an overview of our framework building blocks (see Figure~\ref{fig:encoder_pipeline}). 
First, the main requirement of our method is to have access to a \textit{parametrized augmented reality renderer}. Such a renderer $R$ can render in a source image an object whose appearance is parametrized by a vector of graphics parameters $g$. Thus, for each frame of a video, renderer $R$ takes as input a frame $X$ as well as a vector of parameters $g$ and synthesizes an output frame where the rendered object is realistically inserted. 
Furthermore, we assume that $R$ is not differentiable, as it is the case for most AR rendering pipeline implementations. 
In the context of this paper, we illustrate our approach on two examples of AR renderers for virtual try-on, a makeup renderer (lipstick and eye-shadow), and a hair color renderer, that are illustrated in Figure~\ref{fig:ar_renderers} and described in detail in Section~\ref{sect:ar_renderer}.

Given a particular AR renderer $R$, we propose to build a specialized graphics encoder $E$ that learns to map a single example image into the parameter space of the renderer. 
Our graphics encoder is trained using a two-steps self-supervised approach, which relaxes the need for a labeled training dataset. This makes our framework easily adaptable to many AR rendering applications. 
First, we propose a \textit{differentiable imitator module} $I$ that learns to reproduce the behavior of the non-differentiable renderer $R$ using generative learning. The imitator can then be used as a differentiable surrogate of $R$. 
In order to ensure that the entire parameter space of the renderer is correctly modeled by the imitator network, we introduce a novel parameter sensitivity loss. 
The training procedure of our imitator is illustrated in Figure~\ref{fig:encoder_pipeline} and described in detail in Section~\ref{sect:imitator}.
This method for automatically learning a differentiable renderer could be beneficial various inverse graphics tasks, in particular for renderers using non-differentiable operation such as path-tracing.

Secondly, the learned imitator is used to train the graphics encoder module $E$, by computing a rendering loss in the image space, providing differentiable feedback to optimize the graphics encoder weights. To better constrain our problem, we also use a graphics loss in the space of rendering parameters. This training pipeline is illustrated in Figure~\ref{fig:encoder_pipeline} and described in Section~\ref{sect:encoder}.

 Finally, at inference time, the imitator module is discarded and replaced by the original renderer $R$, which is faster and more portable. Then, given an example image, the encoder $E$ estimates the corresponding vector of parameters that must be passed to $R$ to produce a realistic rendering, as depicted in Figure~\ref{fig:encoder_pipeline}.
 The proposed example based virtual try-on framework can be run in real-time as the rendering parameters $g$ are computed only once at the beginning of the training and then fixed for each frame of the video stream to render. 

\section{Augmented Reality Renderers}\label{sect:ar_renderer}
In this section, we describe the two AR renderers that we choose to illustrate the framework proposed in this paper. We selected two renderers that address popular virtual try-on categories with different rendering principles, physically-based computer graphics for makeup, and pixel statistics manipulation for hair color. 

\subsection{Makeup AR Renderer}
Following similar procedures as in~\cite{li2019lightweight}, we use a graphics-based makeup renderer that takes as input  makeup color and texture parameters and generates realistic images in real-time. 
As shown in Figure~\ref{fig:ar_renderers}, the complete pipeline for lipstick includes a lip mesh prediction model, and an illuminant estimation module. The makeup parameters are manually parametrized for each product.
Specifically, the lipstick rendering is done in two steps: 1) Re-coloring; 2) Texture rendering. For re-coloring, the input lipstick color (R, G, B values) is adjusted to better fit with background illumination based on gray histograms from the input lip. In the second step, texture of the lipsitck (e.g. glossiness, sparkles) is applied using the environment reflection estimation, the estimated lip mesh and a simple material model.
In this paper, as described in Table~\ref{tab:lip_param_desc}, we only consider the main rendering parameters, and leave other material parameters to their default values.

\begin{table}[h!]
    \centering
    \caption{Main parameters used in the makeup rendering process. The complete rendering process includes a total of 17 parameters in order to achieve more sophisticated lipstick looks including sparkles. }
    \small
    \begin{tabular}{cc}
        \hline
        Parameter & Range \\
        \hline
        Makeup opacity & [0, 1]\\
        R,G,B &  [0, 255] \\
        Amount of gloss on the makeup & $[0, +\infty)$ \\
        Gloss Roughness & [0, 1]\\
        Reflection intensity & [0, 1]\\
        \hline
    \end{tabular}
    \label{tab:lip_param_desc}
\end{table}

\subsection{Hair Color AR Renderer}

The hair AR rendering pipeline we use consists of a sequence of image processing primitives combined with hair mask estimation.
First, given a set of swatch parameters (Table~\ref{tab:hair-cms-swatch-parameters}) a re-coloring step computes a pixel-wise color transformation on a reference hair swatch image. From this re-colored swatch, a reference histogram for R, G, B and gray values is extracted.  
The second step in the hair AR rendering pipeline, the rendering process, uses both this swatch histogram and additional non-swatch parameters from the shade matching process to render hair color.
In the rendering process, a hair segmentation model first estimates a hair mask from the input image.
The detected hair region is then transformed so that its histogram matches the histogram of the re-colored swatch image.
Finally, shine and contrast effects boost global and local contrast to improve texture, and a blending effect
can add an optional faded look.
In this paper, we consider both swatch and non-swatch parameters as they both affect all steps of the rendering process.
\begin{table}[t]\centering
	\caption{\label{tab:hair-cms-swatch-parameters}Parameters used in the hair color rendering process.
		The re-coloring step uses swatch parameters to modify the color of a reference swatch image, that is then used as a target for histogram matching.
		Non-swatch parameters affect post histogram-matching steps in the rendering process.
		Blend effect refers to applying a faded look, and the parameter controls the gradient offset from the top of the hair.}
\small
	\begin{tabular}{@{}lc@{}}
		\hline
		Parameter    & Range         \\
		\hline
		\multicolumn{2}{c}{Swatch parameters} \\
		\hline
		Brightness   & $[-0.3, 0.3]$ \\
		Contrast     & $[0.5, 2]$    \\
		Exposure     & $[0.5, 2]$    \\
		Gamma        & $[0.5, 3]$    \\
		Hue          & $[0, 1]$      \\
		Saturation   & $[0, 3]$      \\
		\hline
		\multicolumn{2}{c}{Non-Swatch parameters} \\
		\hline
		Blend Effect & $[-1, 1]$     \\
		Intensity    & $[0, 1]$      \\
		Shine        & $[0, 1]$      \\
		\hline
	\end{tabular}
\end{table}

\section{Learned Imitator Module} \label{sect:imitator}

\begin{figure*}[t!]
\begin{center}
   \includegraphics[width=0.7\linewidth]{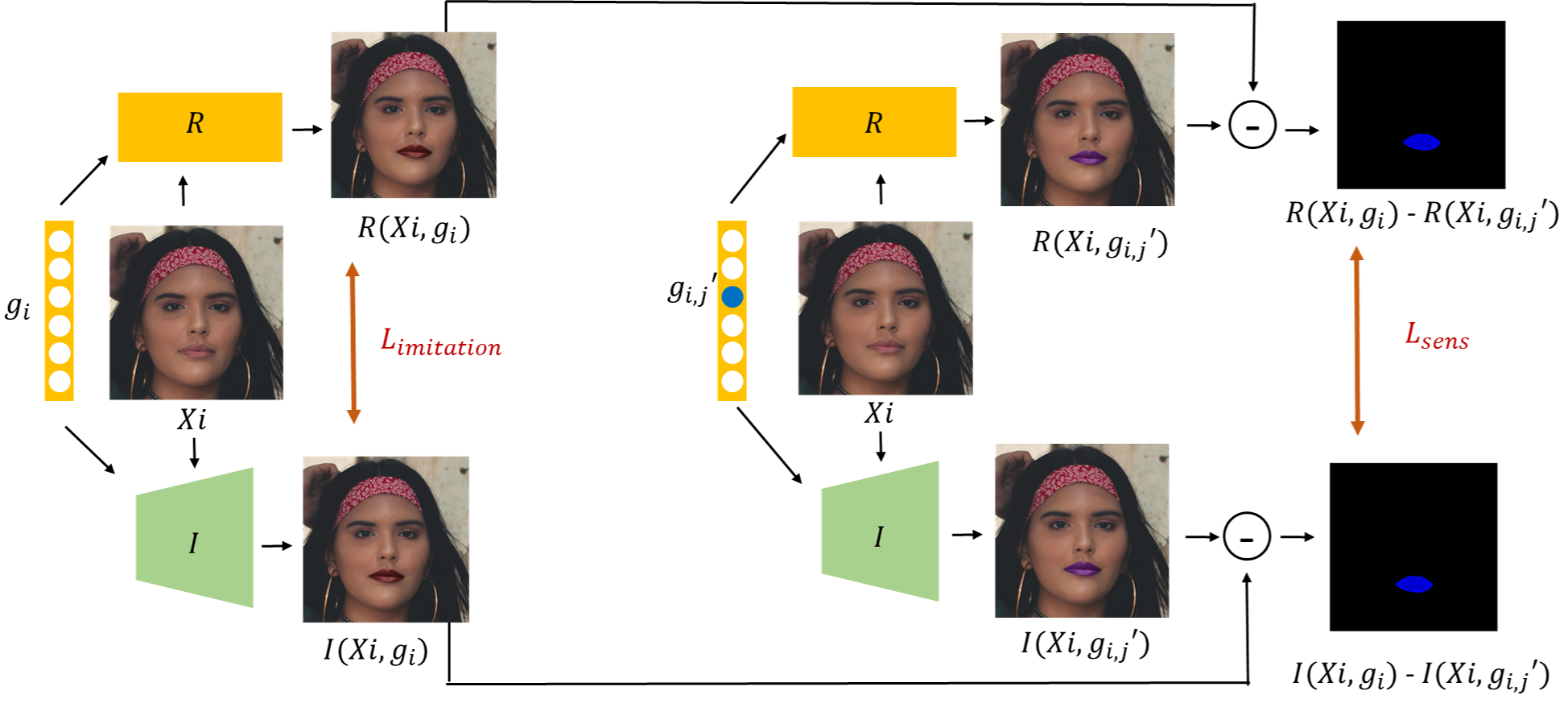}
\end{center}
   \caption{The training procedure of our differentiable imitator module $I$ that learns to reproduce the behavior or the renderer $R$. The imitation loss function $L_{imitation}$ enforces perceptual similarity between $R$ and $I$ outputs on randomly sampled graphics parameter vectors $g$. The sensitivity loss function $L_{sens}$ ensures that a random shift in any dimension of the graphics parameter vector is correctly modeled by the imitator.}
\label{fig:imitator_pipeline}
\end{figure*}

\subsection{Imitator motivation}

Our objective of building an inverse graphics encoder which can be used to perform virtual try-on from a single example image in real-time on mobile device using a conventional, non-differentiable AR-renderer. When training such a model, the ground truth rendering parameters are known during the encoder training and can be used to supervise the encoder directly, as in \cite{kips2021inversemakeup}. However, this assumes that a distance in the space of graphics parameters is a good measure of appearance. This distance can be misleading as some parameters (e.g. RGB for makeup) have a very large contribution to the rendering results, while other parameters (e.g. gloss roughness) have a more limited impact. 
Similarly, reaching a realistic shine effect for a given product often requires very high accuracy in the setting of the shine parameters, where slight variations can lead to perceptually very different textures.
Thus, the graphics loss function, computed in the space of rendering parameters, does not provide an optimal signal for supervising the inverse graphics encoder training. Therefore, we propose to use an additional rendering loss function in the image space, by leveraging an imitator module and a perceptual distance.
In this section, we detail the first step of our framework, which consists in training a \textit{differentiable imitator module} that will later be used to train our inverse graphics encoder. Our imitator takes the form of a generative neural network that automatically learns to reproduce the behavior of a given renderer. This network can then be used as a differentiable surrogate of the initial renderer, for solving various inverse graphics problems. 

The idea of using a generative network to build a differentiable estimate of a renderer was first proposed in the field of automatic avatar creation \cite{Wolf2017, Shi2019, Shi2020}. However, compared to fixed-camera avatar renderers, AR renderers are more complex functions that are usually composed of computer vision and computer graphics modules. Previous renderer imitator methods directly apply the conventional Generative Adversarial Networks method for image-to-image translation established by \cite{Isola2017}. This approach consists in training the imitator to reproduce the renderer output by minimizing a perceptual loss between the renderer and imitator outputs on a set of example rendering images. 
However, this method does not leverage the specificity of the renderer imitation problem, where training phase does not depend on a fixed set of training images, but can be dynamically generated following using the original renderer. 

We propose to leverage this advantage to introduce a more constrained formulation of the imitator problem, based on a novel \textit{rendering sensitivity loss function}. This additional loss term is motivated by two observations. 
First, the imitator network is required to accurately model each of the renderer parameters. However, this is not explicitly enforced by the conventional imitator approach, where parameters that only impact a small portion of the rendered images have a limited weight in the perceptual loss function.
Secondly, in order to accurately solve inverse graphics problems, the imitator needs to learn a continuous representation for each parameter, where a shift in a given parameter 
will lead to changes in the rendered image which are similar to the ones obtained using the actual AR renderer. This is particularly challenging as generative networks are known for the difficulties they encounter in accurately modeling the entire training data distribution (i.e., the mode collapse problem) \cite{liu2019spectral, gulrajani2017improved, arjovsky2017wasserstein}.
Our sensitivity loss, which provides an answer to these problems, is inspired by the expression of the finite difference and 
forces the imitator to learn a rendering function where the derivative with respect to the rendering parameters is the same as the approximated derivative on the non-differentiable renderer. For this reason, this loss does not operate on the image space, as the rendering loss, but in the image difference space. Thus, when sampling rendering parameters, the sensitivity loss forces the generator to modify the same pixels as the non-differentiable renderer, and in the same proportion.

\subsection{Imitator objective functions}

In this section, we detail the training procedure of our imitator model, illustrated in Figure~\ref{fig:imitator_pipeline}. 
Our objective is to train an imitator network $I$ that learns to reproduce the behavior of the renderer $R$, and for which derivatives with respect to $g$ can be computed. We generate training data by randomly sampling $n$ graphics vectors $g_i, i=1...n$ and render them through $R$ with a randomly associated portrait image $X_i$. 
We propose to train $I$ by using a combination of two loss functions. 
First, we use an \textit{imitation loss} function that enforces the imitator network to produce outputs that are perceptually similar to the renderer for a given image $X_i$, and graphics vector $g_i$. The perceptual similarity is computed
using a perceptual distance based on deep features \cite{kim2016accurate}, and  can be written as follows:

\begin{equation}
\label{eqn:perceptual}
L_{perceptual}(x,y) = \left\lVert E_{VGG}(x) - E_{VGG}(y) \right\rVert_2 ^2
\end{equation}
where $E_{VGG}$ is the feature encoder of a pre-trained VGG neural network. Then, our imitation loss function is the following:
\begin{equation}
\label{eqn:imitation}
L_{imitation} = \frac{1}{n}  \sum_{i=1}^n  L_{perceptual} \Big( R(X_i,g_i),I(X_i,g_i) \Big) 
\end{equation}

However, this imitation constraint is not sufficient in practice. We introduce a novel \textit{sensitivity loss function} based on two observations~: (a) the imitator must be able to correctly model all the dimensions of the graphics vector $g$, and (b) the imitator must learn a continuous representation of each dimension of $g$, where a given shift in $g$ will be modeled by changes of the corrected magnitude in the synthesized image.
Our sensitivity loss term enforces additional constraints to satisfy these properties. 
For a given synthesized image $R(X_i,g_i)$ each element $j, j=1...m$ of the graphics vector $g_i$ is randomly sampled independently. Each time, the new sampled vector, noted $g_{i,j}'$, is passed to the imitator and the renderer, 
and the imitator is explicitly constrained to modify the synthesized image in the same proportion as the renderer did.
The sensitivity loss function can be written as follows:
\begin{multline*}
L_{sens} = \frac{1}{n}  \sum_{i=1}^n \sum_{j=1}^{m} \| [R(X_i,g_i) - R(X_i, g_{i,j}')]\\ 
- [I(X_i,g_i) - I(X_i, g_{i,j}')] \|_2 ^2
\end{multline*}


Finally, we use the conventional adversarial GAN loss function, where $D$ is a discriminator module trained with the gradient penalty loss from \cite{gulrajani2017improved} :
\begin{equation}\label{eqn:gan_loss}
L_{GAN} = - \frac{1}{n} \sum_{i=1}^n D(I(X_i,g_i))
\end{equation}

In total, our imitator is trained to minimize the following loss function, where $\lambda_1$ and $\lambda_2$ are weighting factors that are set experimentally:
\begin{equation}\label{eqn:imitator_total}
L^{I}_{total} = \lambda_1 L_{imitation} + \lambda_2 L_{sens} + L_{GAN}
\end{equation}

\section{Inverse Graphics Encoder Module} \label{sect:encoder}


\subsection{Self-supervised training procedure}
Our final objective is to train an inverse graphics encoder network $E$ that learns to map a single reference image to the set of parameters of a given renderer $R$. The training procedure or our inverse graphics encoder module is summarized in Figure~\ref{fig:encoder_pipeline}.

We propose to use a self-supervised training approach where a graphics vector $g$ is randomly sampled and rendered with a random portrait image. Then, we constrain the graphics encoder to recover the original graphics vector from the synthesized image using a \textit{graphics loss} function in the space of rendering parameters: 
\begin{equation}\label{eqn:loss_graphics}
L_{graphics} = \frac{1}{n}  \sum_{i=1}^n \left\lVert g_i - E(R(X_i,g_i)) \right\rVert_2 ^2
\end{equation}

However, using only a graphics loss function suffers from limitations. A distance in the space of rendering parameters might not reflect well a perceptual distance in the image space. For instance, some parameters are central in the rendered image appearance, such as $R, G, B$ values for makeup synthesis, while other parameters will only affect the appearance marginally. Thus, the graphics loss term does not provide an optimal signal for supervising the inverse graphics encoder training. 
Therefore, we propose to use an additional rendering loss function in the image space, by leveraging our imitator module. For each training image, the encoder module estimates a graphics vector, and the imitator computes the corresponding rendered image. A perceptual distance is then calculated between this reconstructed rendering and the original rendered image that was passed to the graphics encoder, as illustrated in Figure~\ref{fig:encoder_pipeline}. The rendering loss can be written as follows:
\begin{equation}\label{eqn:loss_rendering}
L_{rendering} = \frac{1}{n}  \sum_{i=1}^n L_{perceptual} \Big( R(X_i,g_i) - I(X_i, E(R(X_i,g_i))) \Big)
\end{equation}

Since the imitator is differentiable, we can compute the gradients of this loss function with respect to the encoder weights, and train the encoder network using the conventional stochastic gradient descent procedure. In total, our inverse graphics encoder is trained to minimize the following loss, where $\lambda_3$ is a weighting factor to balance the two components of the loss function:
$$ L^{E}_{total} = L_{graphics} + \lambda_3 L_{rendering}$$

\subsection{Encoder Inference}

At inference time, the imitator module is discarded, as it is less optimized for inference on portable devices compared to the initial augmented reality renderer $R$.
Given a single reference image $X_{ref}$, the encoder network directly estimates the set of corresponding rendering parameters $\hat{g}=E(X_{ref})$. 
Thus, for each frame of a video, $R$ can be used to render the virtual try-on in real-time using $\hat{g}$, as illustrated in Figure~\ref{fig:encoder_pipeline}.
Since the rendering parameters are fixed for each frame, the encoder inference needs to be run only once and does not impact the real-time efficiency of the renderer. 


\section{Implementation}
\subsection{Controlling Data Distribution}

\begin{figure}[t!]
\begin{center}
  \includegraphics[width=1.0\linewidth]{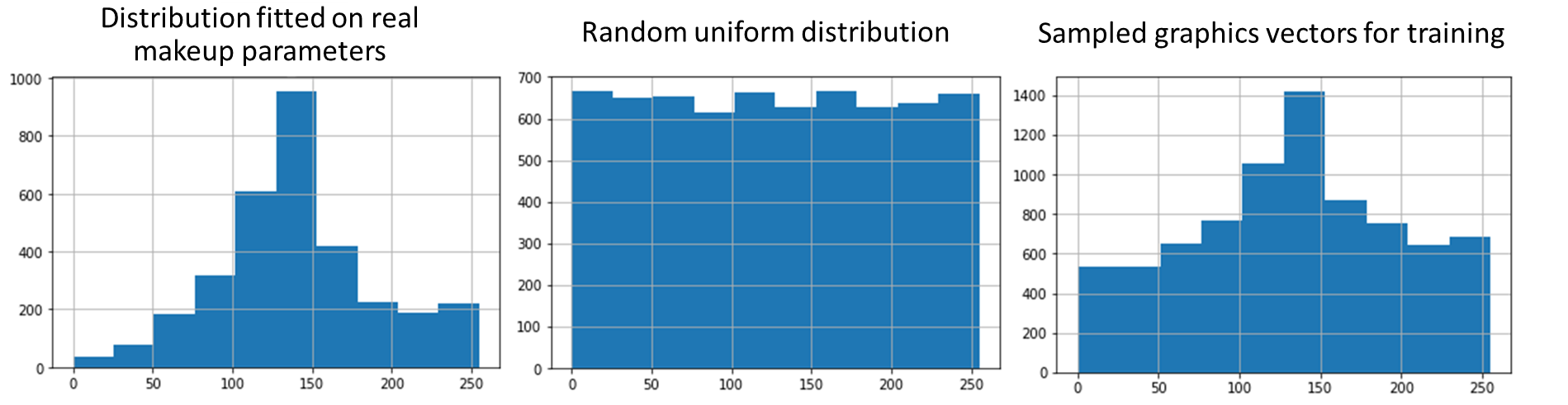}
\end{center}
  \caption{Example of the sampling distribution for the $R$ parameter of our lipstick renderer. Starting from a distribution fitted on natural data, we reinforce the diversity of the training by sampling from a uniform distribution.}
\label{fig:param_distrib}
\end{figure}

A specificity of our framework is that we fully control the distribution of graphics vectors used to train our inverse graphics encoder. We propose to leverage this distinctive trait  to construct a distribution that will reinforce the model performance on extreme examples.
To obtain a realistic graphics parameters sampling, we fit a multivariate normal distribution on a set of rendering parameters previously set by experts to simulate a range of existing products. We choose a Gaussian distribution as it seemed adapted to the empirical distribution of our data.
In addition, we reinforce the diversity of the sampled graphics vectors by alternatively sampling from a uniform distribution as seen in Figure~\ref{fig:param_distrib}
. Even though this might lead to non-realistic synthetic images, the increased diversity will make the framework more robust to extreme examples that might occur in practice, such as blue lipstick or purple hair. 

\subsection{Model Architectures and Training}

\begin{table}[t!]
 \caption{The architecture of our imitator and inverse graphics encoder module}
 \label{tab:model_archi}
 \includegraphics[width=\linewidth]{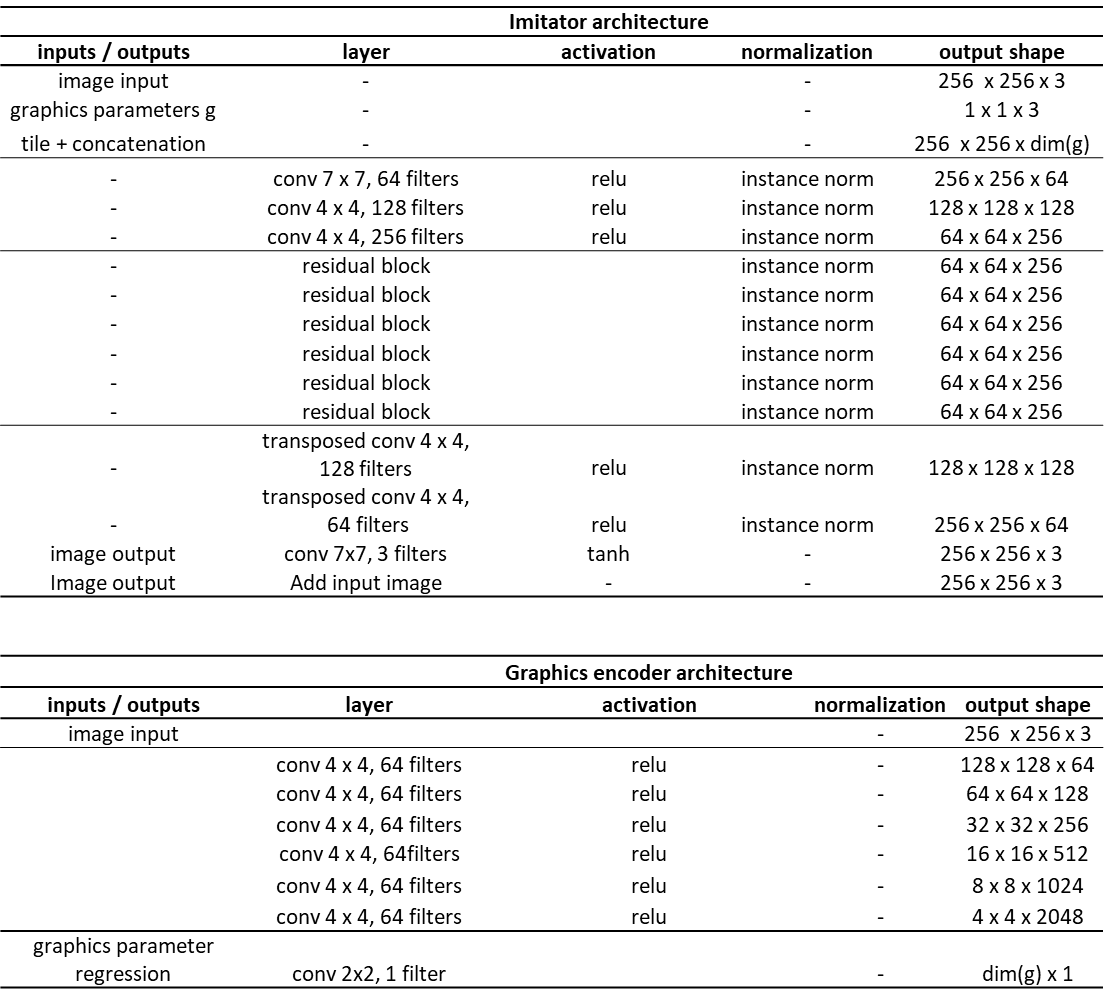}
\end{table}

The architectures of our imitator and inverse graphics encoder module are described in detail in Table~\ref{tab:model_archi}.
Our learned imitator module is implemented using an architecture inspired by the StarGAN \cite{choi2018stargan} generator. The generator input is constructed by concatenating the graphics parameters $g$ to the source image $X_i$ as additional channels.
We use instance normalization in each layer, as well as residual blocks composed of two convolutional layers with $4 \times 4$ kernels and a skip connection. 
Furthermore, in the final layer, the generator outputs a pixel differences map that is added to the source image to obtain the generated image. This architecture allows for a better preservation of the input image details, as the entire image does not need to be encoded in the generator bottleneck. 
The encoder architecture is composed of a simple encoder network based on convolution blocks with $4\times 4$ kernels and ReLU activation outputs. The final layer of the encoder is composed of linear activation outputs of the same size as the graphics vector. 

The training datasets are synthesized by sampling $n=15000$ graphics parameters vectors for lipstick and hair color and rendering them on random portrait images from the \textit{ffhq} dataset \cite{karras2019style}, using renderers described in Section~\ref{sect:ar_renderer}.
To improve the proportion of relevant pixels in the lipstick images, we crop the portraits around the lips before feeding them to the imitator and encoder model.
We empirically set the loss weighting factors values to $\lambda_1=100$, $\lambda_2=1000$ and $\lambda_3=20$ for both experiments.
The imitator and inverse graphics encoder are trained with a batch size of 16 over 300 epochs using an Adam optimizer with a fixed learning rate of $5e-5$.
The code and training data will be released upon acceptance of this paper.

\section{Results}
\subsection{Qualitative Evaluation}
\subsubsection{Imitator}

\begin{figure}[t!]
\begin{center}
\includegraphics[width=0.7\linewidth]{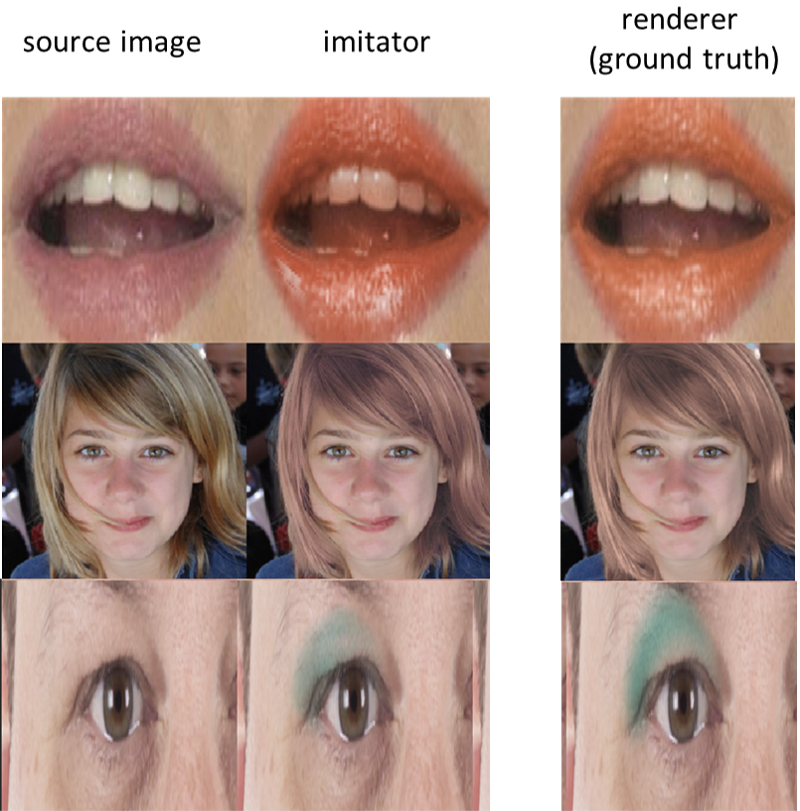}
\end{center}
\caption{Our imitator module learns to accurately reproduce the behavior of complex augmented reality renderers, editing the right pixels and reproducing perceptually similar appearances.}
\label{fig:imitator_example}
\end{figure}

To illustrate the performance of our learned imitator module, we show a qualitative comparison of the imitator $I$ and renderer $R$ outputs in Figure~\ref{fig:imitator_example}. Even though considered AR renders are composed of multiple complex modules for segmentation and computer graphics rendering, our imitator is able to learn how to accurately reproduce their behavior. 
For both renderers, the imitator modifies only relevant image regions, and the material appearance of makeup and hair are correctly reproduced. 

\begin{figure}[t!]
\begin{center}
\includegraphics[width=0.8\linewidth]{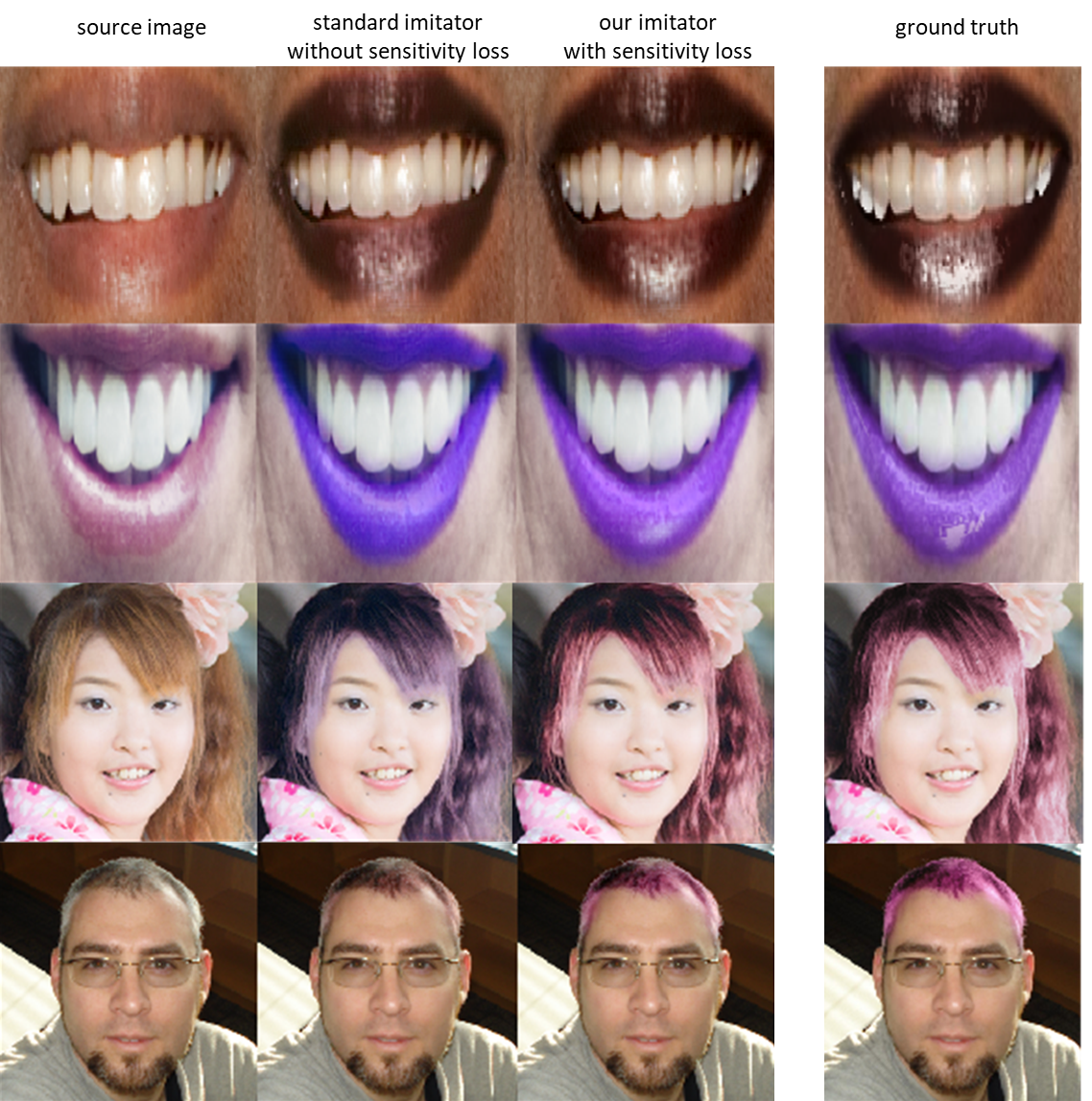}
\end{center}
\caption{A qualitative ablation study illustrates the impact of our novel rendering sensitivity loss. The accuracy in rendering parameters such as shine or color is significantly improved.}
\label{fig:imitator_quali}
\end{figure}

Furthermore, we perform a qualitative ablation study to emphasize the impact of our novel rendering sensitivity loss in Figure~\ref{fig:imitator_quali}. It is worth noting that, compared to the standard imitator approach, our imitator module trained with sensitivity loss learns a better representation for parameters that affect a smaller portion of the image, such as the shine level in the lipstick rendering. It also leads to more accurate rendering of colors in the case of hair color virtual try-on. 

\subsubsection{Graphics encoder}

\begin{figure*}[t!]
\begin{center}
   \includegraphics[width=1.0\linewidth]{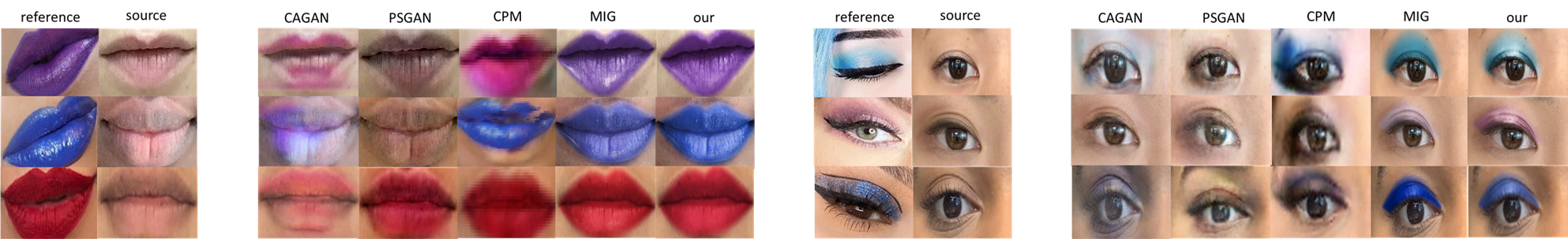}
\end{center}
   \caption{Qualitative comparison of our approach against neural rendering methods for example-based makeup virtual try-on. Our model achieves more realistic results with high resolution. Furthermore, using an imitator allows us to achieve better shine modeling than in MIG}
\label{fig:lips_vs_neural}
\end{figure*}

\begin{figure}[t!]
\begin{center}
  \includegraphics[width=0.9\linewidth]{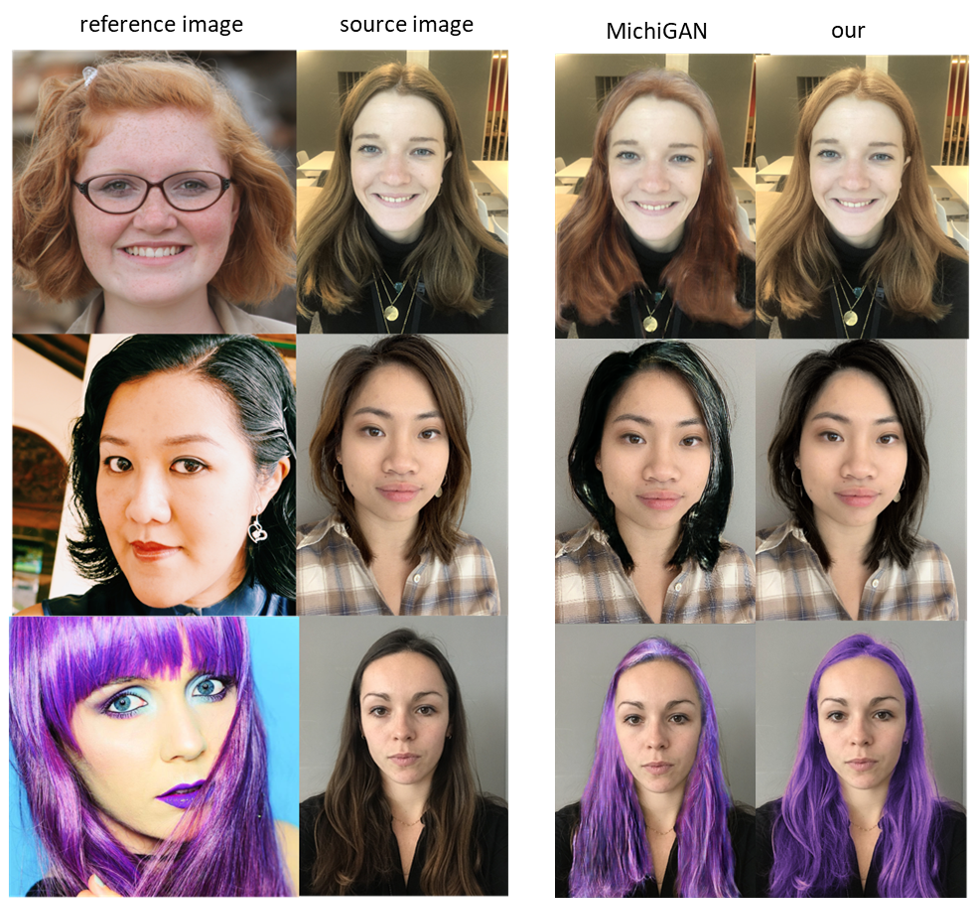}
\end{center}
  \caption{Comparison against generative-based methods for hair color virtual try-on. In addition to reaching real-time on mobile devices, our model produces more realistic hair color results.}
\label{fig:hair_quali}
\end{figure}

We use our inverse graphics encoder to perform virtual try-on from example images, and compare our results against popular neural based models for hair and makeup synthesis. The results are illustrated in Figure~\ref{fig:lips_vs_neural} for makeup and Figure~\ref{fig:hair_quali} for hair. 
The makeup and hair appearances are correctly extracted from the example image and rendered by our framework in high resolution. Furthermore, our encoder is able to model complex appearance attributes, such as lipsticks shiny reflections, and brown hair with blond highlights, as can be seen in Figure~\ref{fig:header_figure}.
Additional video examples are provided as supplementary material.
However, it can be observed that the quality of the results seems lower for the hair color virtual try-on. As it can be seen in Figures~\ref{fig:header_figure} and~\ref{fig:hair_quali}, our framework can reproduce the general hair color, but sometimes fail to accurately model reflection color. 
Compared to other methods, it can be observed that, for makeup, our models achieved more realistic results with higher resolution.
In addition, compared to the MIG \cite{kips2021inversemakeup} our imitator approach allows us to better model the complex textures with shine.
For hair color virtual try-on, we compare our results with MichiGAN, the generative based method from \cite{tan2020michigan}. Our approach produces results that are of comparable realism, while reaching real-time speed on mobile devices, which is not the case for generative models as illustrated in Section \ref{sec:inference_speed}. However, our method is limited to hair color virtual try-on, due to our renderer limitations, while slower generative approaches such as \cite{Patashnik_2021_ICCV} and \cite{xiao2021sketchhairsalon} can also control hair style.

\begin{figure}[t!]
\begin{center}
  \includegraphics[width=1.0\linewidth]{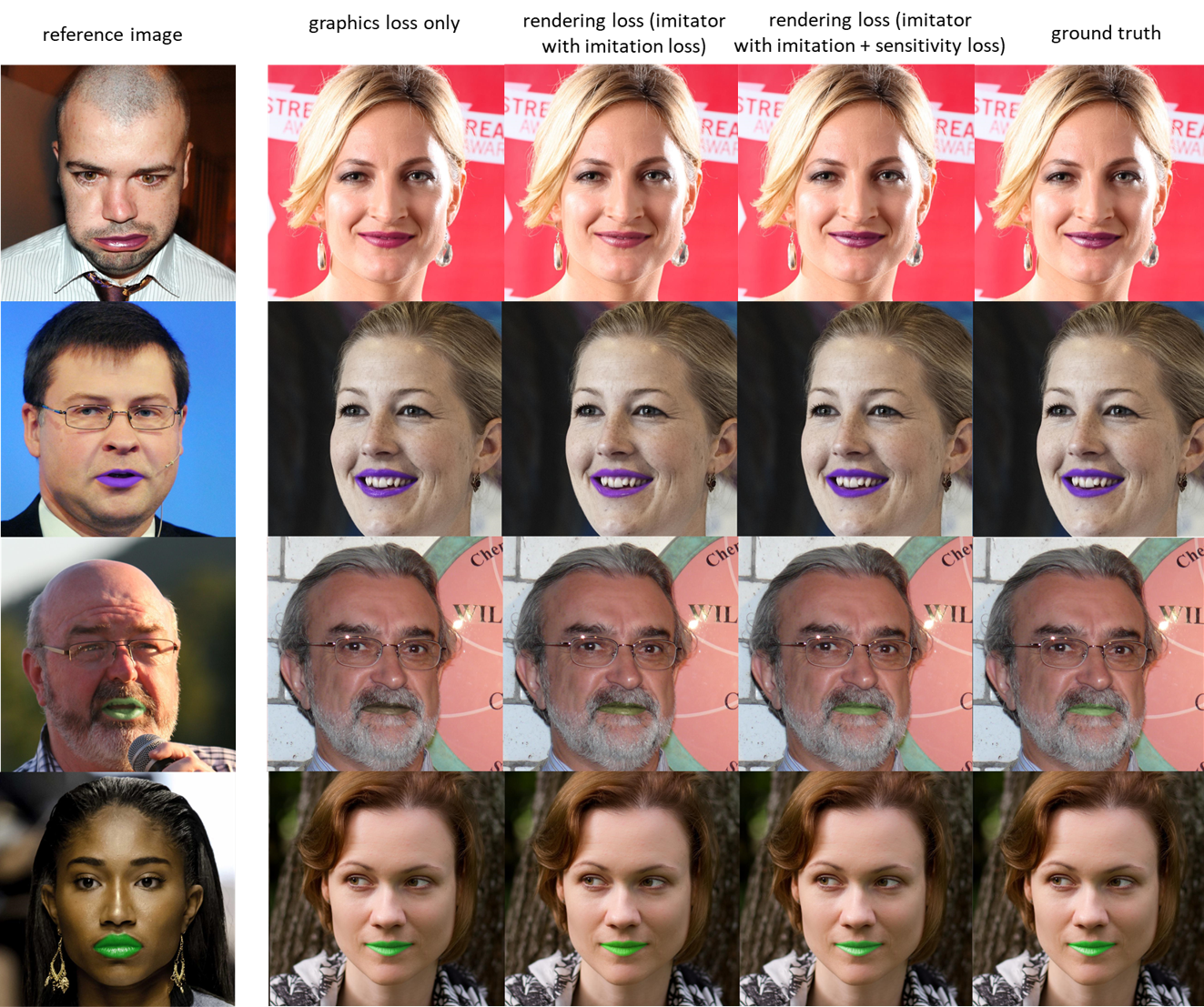}
\end{center}
  \caption{Qualitative ablation study on synthetic data. These results confirm that training the inverse graphics encoder using an imitator based on sensitivity loss leads to improved realism in image based virtual try-on.}
\label{fig:quali_ablation_study}
\end{figure}

\subsection{Quantitative Evaluation}

\subsubsection{Ablation study}
In order to analyze the role played by each component of our framework, we perform a quantitative ablation study. We build a synthetic experiment by sampling 3000 graphics vectors and rendering each of them on two portrait images randomly drawn each time from the \textit{ffhq} dataset, for a total of 3000 different sampled pairs of portrait images of different persons. During the experiment, we extract appearance from the first rendered image using our graphics encoder, and transfer it to the second portrait before rendering.
Ultimately, we compare our estimated rendering with the ground-truth rendering on the second portrait. 
This experiment is illustrated in 
the supplementary materials and the results are reported in Table~\ref{tab:ablation_study_quant}. Several example results for lipsticks virtual try-on are presented in Figure~\ref{fig:quali_ablation_study}.
These results tend to demonstrate the usefulness of using an imitator module, as opposed to a single loss in the space of rendering parameters as in \cite{kips2021inversemakeup}, reducing the average perceptual distance from $0.071$ to $0.054$ for lipsticks.
Furthermore, compared to the standard imitator approach, using our novel rendering sensitivity loss during the imitator training leads to a more accurate inverse graphics encoder.
Lastly, using a combination of graphics loss and rendering loss terms achieves the best performance on most metrics by a small margin, we thus suggest to conserve these two loss terms for training.  



\subsubsection{Evaluation on real makeup data}

To quantitatively compare our method against other image-based virtual-try on methods, we perform an experiment based on real images. 
We reproduce the quantitative evaluation of lipstick virtual try-on from the example introduced in \cite{kips2020gan}. In particular, we use the dataset provided by the authors with 300 triplets
of reference portraits with lipstick, source portraits without makeup, and associated ground-truth images of the same person with the reference lipstick.
We compare our approach against state-of-the-art generative methods for makeup synthesis from example image. The results of this experiment are presented in Table~\ref{tab:quant_mu_transfer}, and confirm that our framework achieves a more realistic virtual try-on according to all metrics, in addition to reaching real-time.
We do not consider an equivalent experiment for hair color, as collecting a similar dataset would require multiple panelists to dye their hair to the same color, which is difficult to achieve in practice. 


\begin{table*}[h]
\caption{Ablation study on lips synthetic data.}
\centering
\small
\begin{tabular}{|c|c|c|c|c|}
\hline
\multicolumn{5}{|c|}{Lipstick experiment} \\
\hline
 Inverse graphics encoder loss & Imitator loss& PSNR $\uparrow$ (mean $\pm$ std) & SSIM $\uparrow$ (mean $\pm$ std) & $\downarrow$ perceptual dist. (mean $\pm$ std)\\
\hline
graphics loss MIG \cite{kips2021inversemakeup} & - & 44.87 $\pm$ 4.92 & 0.997 $\pm$ 0.001  & 0.071 $\pm$ 0.036\\
\hline
rendering loss  & imitation loss & 47.13 $\pm$ 5.02 & 0.998 $\pm$  0.001 & 0.059 $\pm$ 0.033\\
\hline
rendering loss  & imitation loss + sensitivity loss  & 47.85 $\pm$ 5.15 & \textbf{0.999 $\pm$ 0.001} & 0.055 $\pm$ 0.032\\
\hline
graphics loss + rendering loss  & imitation loss + sensitivity loss & \textbf{47.90 $\pm$  5.12} & \textbf{0.999 $\pm$ 0.001} & \textbf{0.054 $\pm$ 0.032}\\
\hline
\multicolumn{5}{|c|}{Hair color experiment} \\
\hline
 Inverse graphics encoder loss & Imitator loss& PSNR $\uparrow$ (mean $\pm$ std) & SSIM $\uparrow$ (mean $\pm$ std) & $\downarrow$ perceptual dist. (mean $\pm$ std)\\
\hline
graphics loss MIG \cite{kips2021inversemakeup} & -  & 28.19 $\pm$ 5.34 & 0.927 $\pm$ 0.056  & 0.797 $\pm$ 0.308\\
\hline
rendering loss  & imitation loss & 26.44 $\pm$ 5.81 & 0.917 $\pm$ 0.068 & 0.810 $\pm$ 0.323 \\
\hline
rendering loss  & imitation loss + sensitivity loss  & 27.49 $\pm$ 5.82 & 0.924 $\pm$ 0.064 & 0.786 $\pm$ 0.325\\
\hline
graphics loss + rendering loss  & imitation loss + sensitivity loss & \textbf{28.73 $\pm$ 5.90} & \textbf{0.933 $\pm$ 0.055} & \textbf{0.756 $\pm$ 0.320}\\
\hline
\end{tabular}
\label{tab:ablation_study_quant}
\end{table*}

\begin{table*}[t]
\small
 \caption{Quantitative evaluation of the makeup transfer performance using a dataset of groundtruth triplet images.}
\centering
\begin{tabular}{|c|c|c|c|}
\hline
 Model & PSNR $\uparrow$ (mean $\pm$ std) & SSIM $\uparrow$ (mean $\pm$ std) & $\downarrow$ perceptual dist. (mean $\pm$ std)\\
\hline
BeautyGAN \cite{li2018beautygan} & 17.44 $\pm$ 3.43 & 0.609 $\pm$ 0.094 & 0.093 $\pm$ 0.018\\
\hline
CA-GAN \cite{kips2020gan}& 17.92 $\pm$ 2.93 & 0.621 $\pm$ 0.033 & 0.077 $\pm$ 0.019\\
\hline
PSGAN \cite{Jiang_2020_CVPR} & 16.11 $\pm$ 2.42 & 0.360 $\pm$ 0.098 & 0.062 $\pm$ 0.018\\
\hline
CPM \cite{m_Nguyen-etal-CVPR21} & 17.87 $\pm$ 3.65 & 0.655 $\pm$ 0.089 & 0.065 $\pm$ 0.022\\
\hline
MIG \cite{kips2021inversemakeup} & 17.82 $\pm$ 2.80 & 0.663 $\pm$ 0.096 & 0.062 $\pm$ 0.016\\
\hline
Ours & \textbf{18.35 $\pm$ 2.63} & \textbf{0.672 $\pm$ 0.100} &  \textbf{0.060 $\pm$ 0.016}\\
\hline
\end{tabular}
\label{tab:quant_mu_transfer}
\end{table*}


\subsection{User Experiment}

We also conduct a user study in order to compare our model to renderer parametrization set by expert artists. We build a validation dataset of 2500 images of volunteers wearing a total of 327 different lipsticks. For each of these lipsticks, artists have carefully set the rendering parameters to reproduce the makeup product appearance. We use 2000 images to estimate the rendering parameters for each of the considered lipstick using our models, computing the median when multiple images per lipstick are available. The remaining 500 images are kept for validation.

Using this dataset, we conduct a user study on six human evaluators. Each of them is presented with an image from the validation set, and the two associated renderings of the same lipstick, based on artists rendering parameters and on parameters estimated by our model. Each rendering image is randomly denoted as rendering A or B to limit bias in the evaluation. Each evaluator must choose among the categories \textit{``both rendering are valid"}, \textit{``only rendering A is valid"}, \textit{``only rendering B is valid"}, and \textit{``both rendering are invalid"}. All images are labeled by three different evaluators. Finally, we removed images for which a majority vote was not reached among the evaluators (19\%). We also removed images where both renderings were considered unrealistic (14\%), assuming this was more due to the renderer limitations than an inaccurate rendering parametrization. The results of this experiment are presented in Table~\ref{tab:user_study}. 

Results indicate that in $48.5\%$ of cases our system outperforms a manual rendering parametrization, while it performs equally in $19.7\%$ of the labeled examples. However, for $31.8\%$ of the images, our system failed to produce a realistic rendering while an artist could manually obtain a convincing result. In particular, our framework seems to fail to correctly model very dark lipsticks, that were not encountered in our training distribution
This study tends to demonstrate that our system can also be used to help artists to create more realistic renderings, by accelerating the currently manual rendering parametrization using example images.

\begin{table}[!]
\caption{Results of our user study comparing our system to manual renderer parametrization made by artists. Each judge is asked to identify which rendering is the most realistic compared to a real reference image.}
\centering
\begin{tabular}{|c|c|}
\hline
Both renderings valid & 19.68\% \\
\hline
Only artists rendering valid & 31.80 \% \\
\hline
Only our rendering valid & 48.52 \% \\
\hline
\end{tabular}
\label{tab:user_study}
\end{table}

\subsection{Inference Speed}
\label{sec:inference_speed}
One of the advantages of our hybrid method combining deep learning and classical computer graphics
is that it does not use neural rendering at inference. Indeed, a commonly recognized challenge of generative methods is that they cannot be deployed for real-time video applications. In this section, we profile and report the inference speed of our inverse graphics encoder and our lipstick renderer.
Our trained models of the inverse graphics encoder and lip detection are converted from TensorFlow to NCNN \cite{ncnn2018framework} and TensorFlow.js to make it runnable on mobile platforms and mobile web browsers. As shown in Table~\ref{tab:profiling}, our method is able to achieve real-time speed even on mobile platforms (iPhone8 Plus, Safari). Furthermore, the slow inference speed of our learned differentiable renderer confirms that current mobile devices hardware does not allow the use of generative networks for real-time video applications in the browser. This reinforces the usefulness of our approach compared to purely generative models such as MichiGAN \cite{tan2020michigan}.

\begin{table}[h]
\small
    \centering
    \caption{Profiling results of our graphics lipstick rendering pipeline on mobile devices in the Safari web browser. To get accurate results, we skip the first 100 frames and average the results of the next 500 frames for each device. 
    }
    \begin{tabular}{| p{15mm} | p{11mm} | p{10mm} | p{12mm} | p{16mm}|}
        \hline
        Device  & Inverse Encoder & Landmarks Detection  & Rendering \& Display & Learned \newline Differentiable Renderer (Imitator) \\
        \hline
        iPhone8 Plus & 26.98ms & 38.50ms & 52.91ms & 835.66ms \\
        iPhoneX & 27ms & 38.46ms & 57.57ms& 841.78ms \\
        \hline
    \end{tabular}
    \label{tab:profiling}
\end{table}

\section{Conclusion}

In this paper, we present a novel framework for real-time virtual try-on from example images. Our method is based on a hybrid approach combining the advantages of neural rendering and classical computer graphics. We proposed to train an inverse graphics encoder module that learns to map an example image to the parameter space of a renderer. The estimated parameters are then passed to the computer graphics renderer module, which can render the extracted appearance in real-time on mobile devices in high resolution.  

Finally, we introduced a learned differentiable imitator modules which relax the need for a differentiable renderer in inverse graphics problem. This imitator approach could be useful to most inverse graphics tasks, in particular when based on renderers using non differentiable operations such as path-tracing. 

Our framework can be easily adapted to other AR renderers since it uses a self-supervised approach that does not require a labeled training set, but only an access to a parametrized renderer.  We illustrated the performance of our framework on two popular virtual try-on categories, makeup and hair color. Furthermore, we believe that our method could be applied to \st{many} other augmented reality problems, in particular when the object of interest is of homogeneous texture and color. Thus, as future work, our framework could be extended to other virtual-try-on categories such as \st{glasses} nail polish or hats.



\bibliographystyle{eg-alpha-doi} 
\bibliography{biblio}       



\end{document}